\documentclass[letterpaper, 10 pt, conference]{ieeeconf}  

\IEEEoverridecommandlockouts

\usepackage{balance} 

\let\labelindent\relax

\usepackage{comment}

\usepackage{amsmath,amssymb} 

\usepackage{mathtools}
\usepackage{array} 
\usepackage{nicefrac}       
\usepackage{breqn}          
\usepackage{textcomp}
\usepackage{bm} 
\usepackage{pifont}
\usepackage[
  separate-uncertainty = true,
  multi-part-units = repeat
]{siunitx}

\usepackage{graphicx}
\usepackage{adjustbox} 
\usepackage{epsfig}

\usepackage{float}
\usepackage{subcaption}
\usepackage[font=small,labelfont=bf]{caption}
\usepackage{lscape}      

\usepackage{tikz}  
\usepackage{makecell}
\usepackage{overpic}
\usepackage{algorithm}
\usepackage[noend]{algpseudocode}

\usepackage{array} 
\usepackage{tabularx}
\usepackage{multirow}
\usepackage{multicol}
\usepackage{booktabs}   
\usepackage{tablefootnote}

\usepackage{paralist}
\usepackage{enumitem}
\setlist[itemize]{noitemsep,leftmargin=*,topsep=0in}
\setlist[enumerate]{noitemsep,leftmargin=*,topsep=0in}

\usepackage{url}            
\PassOptionsToPackage{table}{xcolor}
\usepackage{color,colortbl} 
\usepackage{soul}

\PassOptionsToPackage{capitalize, noabbrev}{cleveref}
\usepackage[pagebackref=false,breaklinks=true,colorlinks,urlcolor=blue,citecolor=blue,linkcolor=blue,bookmarks=false]{hyperref}

\makeatletter
\let\NAT@parse\undefined
\makeatother
\usepackage[numbers,sort&compress]{natbib}

\usepackage{blindtext}
\usepackage{bbm}

\usepackage{lipsum}

\usepackage{titlesec}
\titlespacing{\section}{0pt}{0.3\baselineskip}{0.25\baselineskip}
\titlespacing{\subsection}{0pt}{0.2\baselineskip}{0.15\baselineskip}
\titlespacing{\subsubsection}{0pt}{0.05\baselineskip}{0.03\baselineskip}

\renewcommand{\paragraph}[1]{\vspace{0.2em}\noindent\textit{#1} --}

\usepackage{listings}
\definecolor{codegreen}{rgb}{0,0.6,0}
\definecolor{codegray}{rgb}{0.4,0.4,0.4}
\definecolor{codepurple}{rgb}{0.5,0,0.9}
\definecolor{backcolour}{rgb}{0.95,0.95,0.95}

\definecolor{lightgray}{rgb}{0.9,0.9,0.9}
\definecolor{lightpink}{rgb}{0.98,0.85,0.86}
\definecolor{lightblue}{rgb}{0.68,0.84,0.9}

\lstdefinestyle{mystyle}{
    backgroundcolor=\color{backcolour},   
    commentstyle=\color{codegreen},
    keywordstyle=\color{magenta},
    numberstyle=\tiny\color{codegray},
    stringstyle=\color{codepurple},
    basicstyle=\fontsize{8}{8}\selectfont\ttfamily\ttfamily,
    breakatwhitespace=false,         
    breaklines=true,
    breakindent=0pt,
    captionpos=b,                    
    keepspaces=true,                 
    numbers=none,                    
    numbersep=5pt,                  
    showspaces=false,                
    showstringspaces=false,
    showtabs=false,                  
    tabsize=2
}
\definecolor{color1}{rgb}{.6,.4,.05}
\definecolor{color2}{rgb}{0,.7,.7}
\definecolor{color3}{rgb}{0.35,0.75,0.0}
\definecolor{color4}{rgb}{0.4,0.8,0}
\definecolor{revision_color}{rgb}{1,0,0}

\renewcommand{\comment}[1]{} 

\newcommand{\revisions}[1]{} 

\newcommand{\jeremy}[1]{\textcolor{color2}{\comment{JC: #1}}}

\newcommand{\ag}[1]{\textcolor{color4}{\comment{AG: #1}}}

\newcommand{\modelName}{RoCoDA\xspace}

\title{\LARGE \bf
RoCoDA: Counterfactual Data Augmentation for \\ Data-Efficient Robot Learning from Demonstrations
}

\author{Ezra Ameperosa$^{*1}$, Jeremy A. Collins$^{*1}$, Mrinal Jain$^{1}$, Animesh Garg$^{1, 2}$
\thanks{$^{*}$ equal contribution}%
\thanks{$^{1}$Georgia Institute of Technology, $^{2}$Apptronik, Inc.}%
\thanks{Correspondence to: \href{mailto:eameperosa3@gatech.edu}{eameperosa3@gatech.edu}}%
}

\begin{document}

\maketitle
\thispagestyle{empty}
\pagestyle{empty}


\begin{abstract}

Imitation learning in robotics faces significant challenges in generalization due to the complexity of robotic environments and the high cost of data collection. We introduce \modelName{}, a novel method that unifies the concepts of invariance, equivariance, and causality within a single framework to enhance data augmentation for imitation learning. \modelName{} leverages causal invariance by modifying task-irrelevant subsets of the environment state without affecting the policy's output. Simultaneously, we exploit $SE(3)$ equivariance by applying rigid body transformations to object poses and adjusting corresponding actions to generate synthetic demonstrations. We validate \modelName{} through extensive experiments on five robotic manipulation tasks, demonstrating improvements in policy performance, generalization, and sample efficiency compared to state-of-the-art data augmentation methods. Our policies exhibit robust generalization to unseen object poses, textures, and the presence of distractors. Furthermore, we observe emergent behavior such as re-grasping, indicating policies trained with \modelName{} possess a deeper understanding of task dynamics. By leveraging invariance, equivariance, and causality, \modelName{} provides a principled approach to data augmentation in imitation learning, bridging the gap between geometric symmetries and causal reasoning. Project Page: \href{https://rocoda.github.io}{https://rocoda.github.io}

\end{abstract}

\section{Introduction}
\label{sec:intro}

Recent advances in imitation learning \cite{goyal2023rvt, zhao2023learning, chi2023diffusion, lee2024behavior, mete2024quest} have demonstrated the ability to teach robots complex behaviors by behavior cloning. These methods have shown remarkable success in controlled environments where the training data closely matches the test scenarios. Despite this, current approaches exhibit limited generalization capabilities when deployed in novel environments or on new tasks.


In contrast, fields like natural language processing, computer vision, and audio processing have experienced gains in performance and generalization due to the availability of vast quantities of diverse data scraped from the internet.

Robotics, however, lacks a comparable large-scale, diverse dataset to drive similar breakthroughs. 
Several attempts have been made to collect large-scale robotics datasets \cite{brohan2022rt, vuong2023open, khazatsky2024droid, walke2023bridgedata}, but these efforts remain relatively small in comparison to other subfields of AI.
An additional challenge is that robotics data does not simply contain static observations, but captures a complex dynamic system where states and actions are causally connected. 
However, this additional structure also presents an opportunity. The inherent symmetries and causal relationships between robot actions and environment states offer a rich source of information that can be exploited. 
Recognizing this allows us to focus the model’s attention on task-relevant subsets of the state space and generalize across irrelevant factors.

\begin{figure}[!t]
  \centering
  \includegraphics[width=\linewidth]{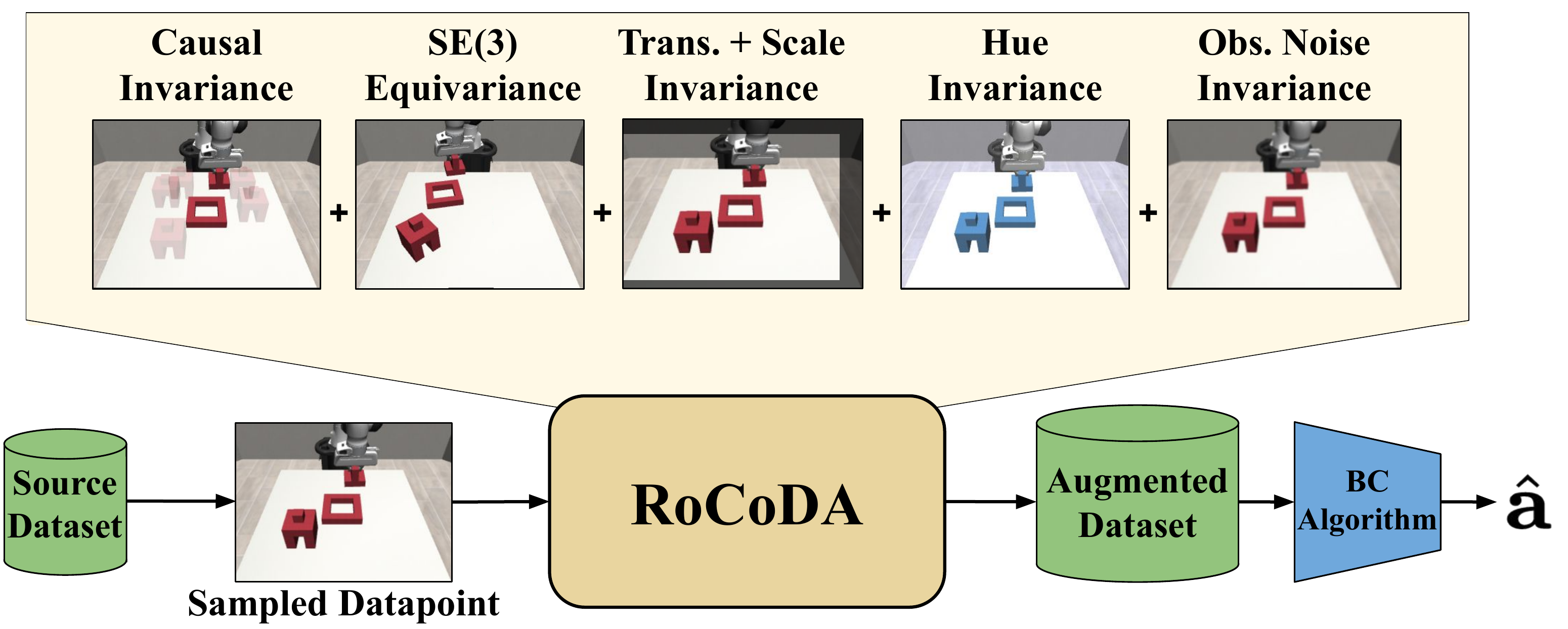}
  \caption{\modelName is a data augmentation framework that leverages causality and several symmetry groups to expand BC datasets. Our pipeline consists of 5 augmentations applied in tandem -- causally invariant counterfactual data augmentation, $SE(3)$ equivariant object state and trajectory augmentation, translation and scale invariance (resize and crop), hue invariance (color jitter), and observation noise invariance (proprioception noise).}
  \label{fig:headliner}
\vspace{-3mm}
\end{figure}
Our key insight is that the causal relationships inherent in robotics tasks can be harnessed to create more robust, scalable policies. By understanding which components of the task are causally dependent, invariant, or equivariant to the robot’s actions, we develop a framework for augmenting training data to enable policy generalization and robustness.

We present Counterfactual Data Augmentation for Robot Learning (\modelName). Our contributions are as follows:
\begin{enumerate}
    \item \textbf{Unified Framework for Data Augmentation:} We formalize the relationship between invariance, equivariance, and causality using group theory and probabilistic graphical models. This framework allows \modelName{} to exploit geometric symmetries and causal structures to produce robust and generalizable policies for complex robotic tasks.
    
    \item \textbf{\modelName:} We introduce a novel data augmentation method that leverages causal invariance, $SE(3)$ equivariance, and visual invariance to improve the robustness, generalization, and performance of behavior cloning policies.

\end{enumerate}

\section{Related Work}

\paragraph{\textbf{Behavior Cloning}}
Behavior Cloning (BC) is a fundamental approach in imitation learning where a policy is trained to mimic expert demonstrations by directly mapping states to actions. Several recent BC methods have enabled robots to execute complex tasks by mimicking teleoperation data. \cite{goyal2023rvt, shridhar2023perceiver} classify 3D subgoals for an end-effector to achieve, and \cite{zhao2023learning, wang2023affordance}  predict latent actions using a CVAE, while \cite{chi2023diffusion,lee2024behavior, mete2024quest} use diffusion, VQ-VAE, and causal convolution, respectively. These methods have proven extremely effective at fitting fine-grained, multi-modal action distributions in their training data. 

\jeremy{TODO: ground this with a couple of specific, cited examples of brittleness in BC algorithms}
Despite their effectiveness on data similar to that in their datasets, BC methods are typically sensitive to distributional shifts, resulting in decreased performance when encountering unseen states or variations in the environment. Our approach addresses this limitation by incorporating causal, invariant, and equivariant augmentations, improving the policy's robustness and generalization capabilities.

\paragraph{\textbf{Data Augmentation for Robot Policy Learning}}
%
%
Data augmentation improves how well robot policies perform, especially when there is limited data or challenges like moving from simulation to the real world. Past work shows that techniques like random crops on images or changing simulation settings help improve policy performance. Random cropping in image observations has shown improvements for both RL and BC methods~\cite{laskin2020RLwithAugData, young2020VIeasy, mandlekar2023mimicgen, kostrikov2020augallyouneed, mandlekar2021matters,chi2023diffusion}.
Domain randomization has been employed to bridge the sim-to-real gap by varying simulation parameters (e.g., texture, friction, lighting) for reinforcement learning algorithms \cite{akkaya2019solving}. State-based augmentations, such as adding Gaussian noise \cite{sinha2022s4rl} improve policy resilience to environmental changes by smoothing the learned state-action space. \cite{mandlekar2023mimicgen} applied 3D transformations to target object poses and corresponding actions, thereby ensuring that the policy is equivariant to these transformations. Recently, generative models have been used for synthetic data generation as well~\cite{chen2023genaug, yu2023scaling,lu2024synthetic}. 

\paragraph{\textbf{Causal Data Augmentation}}
%
Causal data augmentation has been well explored in the context of reinforcement learning \cite{pitis2020counterfactual, pitis2022mocoda, lu2020sample}. CoDA~\cite{pitis2020counterfactual} introduces counterfactual data augmentation for reinforcement learning by resampling subsets of the state space while preserving causal dependencies. MoCoDA \cite{pitis2022mocoda} extends CoDA by incorporating a learned dynamics model, allowing for the handling of overlapping parent sets and enhancing the generation of causally consistent data for RL in more complex environments. 
Both of these methods focused on sample efficiency in low-dimensional tasks for offline RL. In contrast, our work enables robustness and generalization for vision-based behavior cloning algorithms on a set of complex, multi-step tasks.

Our method, \modelName{}, builds upon these techniques by combining geometric equivariance with causal invariance to generate more diverse and causally consistent training data, thereby improving efficiency and generalization. 


\section{A Unified Perspective on Data Generation}
\label{sec:method}

In this section, we present a unified framework that connects the concepts of invariance, equivariance, and causality, which are central to our method, \modelName{}.

The notions of invariance and equivariance are fundamental in understanding the relationship between observation and action in robotics. These concepts are rooted in group theory, which provides a mathematical framework for modeling symmetries and transformations.

\paragraph{\textbf{Groups and Symmetries}}
A group $G$ is a set equipped with an operation that satisfies four properties: closure, associativity, identity, and invertibility. Groups are essential for modeling symmetries and transformations in various domains, including robotics.





\paragraph{\textbf{Invariance and Equivariance}}
%
The function $f: X \rightarrow Y$ is \textit{\( G \)}-equivariant if $\forall$ \( x \in X \) and \( g \in G \), $f(g \cdot x) = g \cdot f(x)$,
where \( \cdot \) denotes the group action of \( G \) on the respective spaces. Equivariance ensures that applying a group transformation before or after \( f \) yields consistent results under the group action on \( Y \).

\noindent \( f \) is \textit{\( G \)-invariant} if, $\forall$ \( x \in X \) and \( g \in G \), $f(g \cdot x) = f(x)$.
Invariance is a special case of equivariance where the group action on \( Y \) is trivial, i.e. \( g \cdot y = y \) $\forall$ \( y \in Y \) and \( g \in G \).





\begin{figure}[!t]
  \centering
  \vspace{-1mm}
  \includegraphics[width=\linewidth]{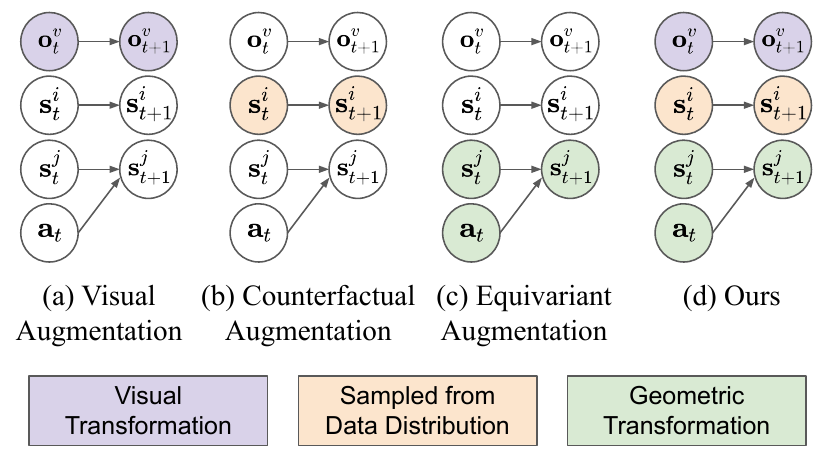}
  \caption{Causal structures for various augmentation methods. (a) Visual augmentation performs transformations to images that do not affect state-action dynamics in the environment. (b) Counterfactual augmentation factors the state space into causally invariant subspaces ($s^i \text{ and } s^j$), and resamples these subspaces from the training data distribution to generate synthetic data. (c) Equivariant augmentation performs a transformation on the pose of a target object associated with a subtask, and uses that same transform to augment the associated actions. (d) \modelName{} unifies these augmentation schemas.}
  \label{fig:pgm}
   \vspace{-3mm}
\end{figure}

\begin{figure*}[!t]
  \centering
  \includegraphics[width=0.97\linewidth]{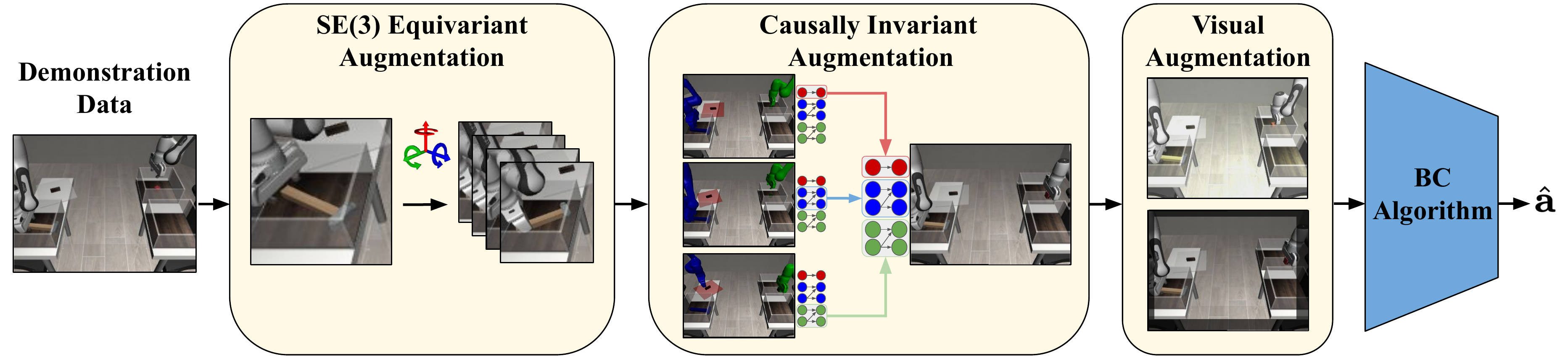}
  \vspace{-2mm}
  \caption{\modelName consists of three stages: $SE(3)$ equivariant state-action augmentation, causal augmentation, and visual augmentation. We first apply rigid body transformations to object poses and identically transform the corresponding actions. Then, we resample subsets of the environment state that are causally invariant to the action. Lastly, we apply several standard augmentations, including random resize/crop, color jitter, and state noise, all of which leverage invariance aspects of the observation with respect to actions (see Section \ref{sec:standardaug}).
  \ag{what is classical augmentation. not the best choice of words. also use not RGB as colors in the grasp. the graph is way to small to be understood.}}
  \label{fig:method}
   \vspace{-4mm}
\end{figure*}

\paragraph{\textbf{Policies as Group Actions}} A policy $\pi: \mathcal{S} \rightarrow \mathcal{A}$ represents a group action when the following conditions hold:
\begin{enumerate}
    \item There exists a group $G$ that acts on the state space $\mathcal{S}$ via a group action $\phi_{\mathcal{S}}: G \times \mathcal{S} \rightarrow \mathcal{S}$. This means that for any group element $g \in G$ and state $s \in$ $\mathcal{S}$, the transformation $g \cdot s$ is a valid operation in $\mathcal{S}$.
    \item There is a corresponding action of $G$ on the action space $\mathcal{A}$, denoted $\phi_{\mathcal{A}}: G \times \mathcal{A} \rightarrow \mathcal{A}$. This ensures that applying a group transformation to the state will induce a transformation on the action.
    \item The policy $\pi$ is \textit{equivariant} with respect to the group action, meaning that applying a transformation $g \in G$ to the state induces the same transformation in the action:
\end{enumerate}
$$\pi(g \cdot s)=g \cdot \pi(s), \quad \forall g \in G, \forall s \in \mathcal{S}$$

This relationship implies that the policy respects the group structure and that actions taken in transformed states are consistent with the transformations applied to the original states.
We leverage \textit{geometric equivariance} by applying random $SE(3)$ (rigid body) transformations to object poses and transforming the corresponding actions, thus generating diverse yet consistent training examples that respect the task's spatial symmetries.

\subsection{\textbf{Causality: Data Generation with Counterfactuals}}


The state input $s$ to a robot policy $\pi(s)$ can often be decomposed into partitions $s=\left(s_C, s_I\right)$, where $s_C$ contains causally relevant variables, and $s_I$ contains causally irrelevant variables with respect to the action $a$.
We can then consider the policy as a function of $s_C$ alone: $\pi(s)=\pi\left(s_C, s_I\right) =\pi\left(s_C\right)$, 
implying that the action is \textit{invariant} to changes in $s_I$.
Notably, this independence of $s_C, s_I$  allows us to manipulate $s_I$ without affecting the action $\pi(s)$. Subsequently, we can generate \textit{counterfactual} data points,  by choosing different $s_I$, without affecting $\pi(s)$, as long as $s_C$ remains unchanged.

We can say that the policy $\pi(s)$ is \textit{causally invariant} to $s_C$. In this context, the policy represents a function with some group action $G$ consisting of transformations that only affect causally irrelevant components of the state, while leaving causally relevant components unchanged:
\vspace{-2mm}
$$\pi\left(g \cdot\left(s_C, s_I\right)\right)=\pi \left(s_C, g \cdot s_I\right)=\pi\left(s_C\right), \quad \forall g \in G$$

\paragraph{\textbf{Factored MDPs}}
We model the environment as a Markov Decision Process (MDP) defined by the tuple $\langle\mathcal{S}, \mathcal{A}, P\rangle$, where $\mathcal{S}$ is the state space, $\mathcal{A}$ is the action space, and $P$ is the transition function.
In complex environments with multiple objects, the number of states can be exponential in the number of objects, making learning challenging \cite{pitis2022mocoda}. To manage this complexity, we consider Factored MDPs (FMDPs), where the state and action spaces are described by sets of variables $\left\{X^i\right\}$, such that $\mathcal{S} \times \mathcal{A}=\mathcal{X}^1 \times \mathcal{X}^2 \times \cdots \times \mathcal{X}^n$. Each state variable $X^i$ depends on a subset of variables $\mathrm{Pa}\left(X^i\right)$ at the previous time step, known as its parents.

\paragraph{\textbf{Local Causal Models}}
In practice, the global factorization assumed by FMDPs is rare. We instead leverage Local Causal Models (LCMs), where the state-action space decomposes into disjoint local neighborhoods $\left\{\mathcal{L}_k\right\}$. Each neighborhood $\mathcal{L}_k$ is associated with its own transition function $P^{\mathcal{L}_k}$ and causal graph $\mathcal{G}^{\mathcal{L}_k}$.
In this framework, if $\left(s_t, a_t\right) \in \mathcal{L}_k$, each state variable $X_{t+1}^i$ depends on its parents $\mathrm{Pa}{ }^{\mathcal{L}_k}\left(X_{t+1}^i\right)$ at the previous time step: $X_{t+1}^i \sim P_i^{\mathcal{L}_k}\left(\mathrm{~Pa}^{\mathcal{L}_k}\left(X_{t+1}^i\right)\right)$. We refer to the tuple $\left\langle X^i, \mathrm{~Pa}^{\mathcal{L}_k}\left(X^i\right), P_i^{\mathcal{L}_k}\right\rangle$ as \textit{a causal mechanism}.

This local factorization captures the idea that objects may interact over time but do so in a locally sparse manner. For example, during a dual-arm handover task, one arm's actions directly affect an object and the other arm only during the handover causal phase.

A \textit{causal graph} is a directed acyclic graphical model where nodes represent variables (state components), and edges represent causal dependencies. The adjacency matrix $A$ of the graph encodes these dependencies, where $A_{i j}=1$ indicates that $X^j$ is causally dependent on $X^i$.
By constructing causal graphs for each causal phase (subtask), we can determine the local causal structures governing the dynamics, thereby enabling us to partition the state into causally independent subsets, enabling efficient data augmentation through independent sampling from these partitions.

\section{\modelName: Generalized Data Augmentation}

\modelName aims to improve the generalization and robustness of imitation learning policies by leveraging both the causal structure inherent to robotic tasks and the equivariance of robotic actions to task-relevant objects. A graphical model of our method vs other data augmentation methods is shown in Figure \ref{fig:pgm}. Notably, previous works only perform a single type of augmentation, while we can perform fundamentals-guided simultaneous augmentations across multiple independently varying quantities.

We achieve this by augmenting a small dataset in the following manner:

\begin{enumerate}
    \item \textbf{Counterfactual Data Augmentation:} We create a causal graph for each subtask within the demonstrations. This allows us to resample and mix subsets of the state space from different trajectories, ensuring causal consistency.
    \item \textbf{Equivariant Data Augmentation via $SE(3)$ Transformations:} We expand the dataset by exploiting the special Euclidean group ($SE(3)$) equivariance of actions with respect to object poses. This involves applying random transformations (translation + rotation) to object poses and adjusting the corresponding actions accordingly. 
    \item \textbf{Standard Augmentation:} We also apply standard data augmentations, which we refer to as Camera Translation Invariance (random crop), Hue Invariance (color jitter), and Observation Perturbation Invariance (state noise), to enhance the diversity of the dataset.
\end{enumerate}

\subsection{Counterfactual Data Augmentation}
\label{sec:CoDAaug}
We leverage the causal relationships between different entities in the environment to augment our dataset. This involves generating causal graphs for each \textit{causal phase} (subtask) and resampling subsets of the state space while maintaining causal consistency.

For a particular demonstration, let the state of the $i$-th entity be denoted as $X_i$. The state for the entire system can be represented as $\mathbf{X}=(X_1, X_2, \dots, X_n)$ where $n$ is the number of entities in the environment. If the entities are causally independent, we can assume that the joint distribution of their states can be factored into independent marginal distributions:
\vspace{-2mm}
$$P(\mathbf{X})=P(X_1,\dots, X_n)=\prod_{i=1}^{n}P(X_{i})$$
Similarly, if causal dependencies exist between $X_1$ and $X_2$ and are independent of all other entities, the joint distribution can be factorized as independent subsets of states:
\vspace{-2mm}
$$P(\mathbf{X})=P(X_1,X_2)\prod_{i=3}^{n}P(X_{i})$$
Notably, the state can be factored into independent subsets, the data can be augmented by sampling in the product space of individual subsets~\cite{pitis2020counterfactual}. In the context of robotic manipulation, sampling from the product space of factorized state partitions involves selecting object states from different demonstrations while ensuring consistency with the joint distribution of dependent states.

Moreover, we define \textit{causal phases} in a task trajectory with a heuristic based on invariant transitions. For manipulator arms, changes in the gripper state (i.e., when the gripper opens or closes) result in non-interacting causal phases in the task. 
This segmentation allows us to construct causal graphs for each subtask. 
We can define causal graphs for each causal phase, by identifying causally independent entities during each phase.
For tasks where multiple agents are present, it is often not practical to define a causal graph for all combinations of causal phases for all agents. Hence, we generate individual causal graphs for each agent, for each causal phase. 

For each causal phase, we construct an adjacency matrix using the dependencies of state variables across time. We construct a joint causal graph over all state variables as a joint block-diagonal adjacency matrix.
Let $A_1$ and $A_2$ be the adjacency matrices for two agents.
We compute this joint adjacency matrix as:
\vspace{-2mm}
$$A = (A_1 \vee A_2)\vee(A_1 \vee A_2)^T,$$

The aforementioned process partitions the state space into subsets given the causal graph(s) associated with a causal phase. Entities that are causally dependent are placed in the same partition of the state space. For each partition, we sample states from a \textit{different trajectory} in the dataset in the \textit{same causal phase}, ensuring that causally dependent entities remain consistent. Independently sampling states from each group creates new states that respect the causal structure of the causal phase. By maintaining causal dependencies, these new states result in valid trajectories that the policy can learn from. 
In practice, we also use a simulator to sample object states from partitions of the state space, and then render the resulting augmented state before inputting it into an image-conditioned behavior cloning algorithm.

Counterfactual data augmentation creates synthetic datapoints which are \textit{out-of-distribution} but follow the same causal structure as data generating distribution.


\subsection{$SE(3)$-Equivariant Data Augmentation}

We exploit the $SE(3)$ equivariance of robotic actions to object poses. This means that applying a transformation to the object poses can be compensated by a corresponding transformation to the actions, preserving the structure of the task. Formally, let $T_{obj}(s_t)$ be the homogeneous transformation between the pose of a target object in the source dataset and the current pose of the target object, and let $T_{act}(a_t)$ be the corresponding transformation applied to the current action. Then we have 
$s' = T_{obj}(s_t)$, and $a' = T_{act}(a_t)$.

Let $D$ be a limited dataset of task demonstrations $D = \{\tau_i\}_{i=1}^{N}$, where each trajectory $\tau_i = {(s_t, a_t)}_{t=1}^{T_i}$ consists of states $s_t$ and actions $a_t$. These trajectories are split into subtasks to generate sub-trajectories, with each subtask being associated with a target object. Sub-trajectories are augmented by randomly generating an unseen pose for the target object, sampling a sub-trajectory from the original dataset, and transforming this sub-trajectory using the homogeneous transformation between the target object and the same object's pose in the original dataset. This preserves the gripper pose from the original dataset with respect to the target object at every time step. Following \cite{mandlekar2023mimicgen}, we add a linear interpolation segment at the start of the augmented sub-trajectory to move the robot into a path that overlaps with the corresponding gripper pose in the sampled original segment. We additionally save states and actions from augmented sub-trajectories that successfully completed the subtask, and discard those that did not succeed.

We assume that we can determine whether an augmented trajectory successfully completes a subtask, in line with prior work~\cite{mandlekar2023mimicgen}. Trajectories that do not result in successful completion are discarded, ensuring that the policy learns from valid demonstrations.

\subsection{Standard Data Augmentation}
\label{sec:standardaug}

Following prior work, we apply traditional data augmentation techniques to the observations to enhance robustness:

\paragraph{Camera Translation Invariance}
Random cropping has found success in training image-based methods \cite{laskin2020RLwithAugData, young2020VIeasy, mandlekar2023mimicgen, chi2023diffusion, kostrikov2020augallyouneed, mandlekar2021matters}. 
We use random resizing and cropping to encourage translation and scale invariance.

\paragraph{Lighting and Hue Invariance}
We use color jitter to make the model robust to lighting conditions and object colors. Note that this cannot be used when colors are relevant to the task, such as in the \textit{Stack Three} task.

\paragraph{Observation Noise Invariance}
Inspired by \cite{sinha2022s4rl}, we find that adding Gaussian noise to observation enhances performance and improves model robustness.



\begin{table*}[!t]
\centering
\begin{minipage}{0.6\textwidth}
\centering
\scriptsize 
\resizebox{\textwidth}{!}{
    
\begin{tabular}{l|c|c|c} 
    \toprule
    \rowcolor[HTML]{CBCEFB} 
    &  Three Block Stack & Three Piece Assembly & Coffee\\ 
    \midrule    
    \rowcolor[HTML]{EFEFEF} 
    No Augmentation (Vanilla-ACT) &  0.33\,\textpm\,0.47&  0.33\,\textpm\,0.47&0.0\,\textpm\,0.0\\ 
    MimicGen \cite{mandlekar2023mimicgen} (ACT) &  57.0\,\textpm\,3.30 & 9.3\,\textpm\,0.94& 44.3\,\textpm\,1.70\\ 
    \rowcolor[HTML]{EFEFEF} 
    
    \modelName w/o \{Visual, Causal\} & 58.7\,\textpm\,3.09 & 11.3\,\textpm\,0.47 & 45.0\,\textpm\,1.63\\ 
    \modelName w/o \{Visual\}  & 69.3\,\textpm\,1.25 & 13.7\,\textpm\,1.89 & 47.0\,\textpm\,2.16\\ 
    \rowcolor[HTML]{EFEFEF}
    
    \textbf{\modelName} (Full)  & \textbf{71.3\,\textpm\,1.25} & \textbf{15.7\,\textpm\,2.05}  &  \textbf{49.3\,\textpm\,1.25} \\

    \bottomrule
\end{tabular}}
\end{minipage}
\hspace{20pt}
\begin{minipage}{0.29\textwidth}
\centering
\caption{Success rate (\%) We compare the maximum success rate over policy rollouts and average over 3 seeds of \modelName{} to standard data augmentation techniques, counterfactual data augmentation and equivariant state-action augmentation on separate rollouts.} 

\label{tab:main}    
\end{minipage}
\vspace{-2mm}
\end{table*}

\section{Results}

Through our experiments, we aim to answer the following questions:

\noindent\textbf{RQ1}: How does encoding causal dependencies between states and actions enhance policy performance compared to models that assume i.i.d. data?

\noindent\textbf{RQ2}: What is the impact of counterfactual data augmentation on the generalization capabilities of behavior cloning algorithms?

\noindent\textbf{RQ3}: Does counterfactual data augmentation improve sample efficiency in policy learning compared to existing approaches?





\subsection{Experimental Setup}

We evaluate our method, \modelName{}, on a variety of multi-step robotic tasks using Action Chunking Transformer (ACT)~\cite{zhao2023learning} as our base imitation learning algorithm. ACT is an encoder-decoder transformer model that learns to autoregressively output latent actions generated by a conditional variational autoencoder (CVAE).

\noindent \textit{Remark.} Although prior work has shown that multi-task imitation learning can be achieved by scaling data and selecting appropriate algorithms \cite{brohan2022rt, brohan2023rt, vuong2023open, bharadhwaj2024roboagent}, we focus on the single-task setting for each model in this work. This enables us to isolate the effects of our data augmentation algorithm on individual tasks without confounding factors introduced by multi-task learning.

\begin{figure*}
  \centering
  \includegraphics[width=0.95
  \linewidth]{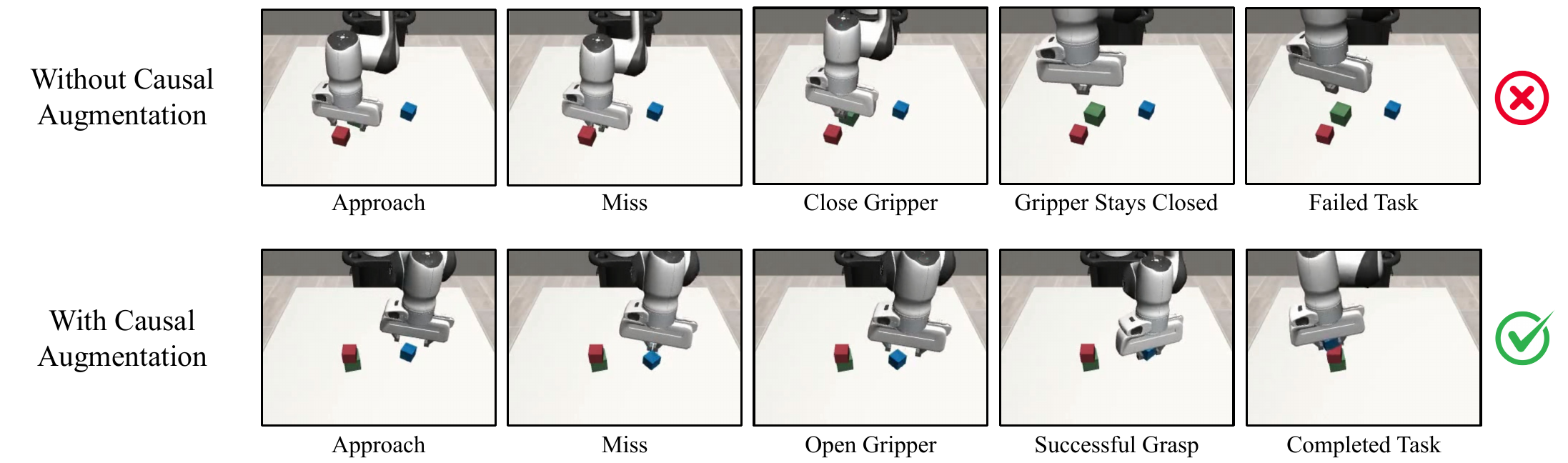}
  \vspace{-2mm}
  \caption{\modelName learns to re-grasp cubes in the \textit{Three Block Stack} task, despite this behavior not being present in the dataset. This suggests that policies trained with \modelName{} possess a deeper understanding of the causal structure underlying the task, enabling the robot to adapt to out-of-distribution inputs.}
  \label{fig:case_study}
   \vspace{-2mm}
\end{figure*}




\paragraph{\textbf{1. Tasks}} We evaluate our method on the following tasks:

\noindent \textbf{Three Block Stack:} The robot must stack three distinct colored blocks in a predefined sequence based on their color. 

\noindent \textbf{Three-Piece Assembly:} The robot assembles three pieces by stacking them in a specific orientation.

\noindent \textbf{Coffee Task:} The task involves placing a coffee pod into a coffee machine and closing the lid.

\noindent \textbf{Transport:} This task involves two robotic arms. The left arm removes a lid and picks up a hammer, while the right arm moves a cube from a back bin to a front bin, takes the hammer from the left arm, and places it in another bin. 

\noindent \textbf{Libero-Object:} We use a task from the Libero-Object dataset where the robot must pick up a specific object (e.g., tomato sauce) and place it in a basket. 

\ag{the details of the task perhaps can be put on the website. we can have a shorter version here}

\vspace{-3mm}

\paragraph{\textbf{2. Baselines}} 
We use ACT, a popular BC baseline \cite{zhao2023learning}. We denote ACT vanilla as trained on unaugmented datasets. 
Further, we use a state-of-the-art augmentation schema MimicGen~\cite{mandlekar2023mimicgen}, that uses $SE(3)$ transformations for data augmentation. We further augment the data with random cropping, as was done in the original work.  



\subsection{Experiments}
\label{sec:experiments}
\paragraph{\textbf{1. Policy Performance}}%
To address \textbf{RQ1}, we compare the performance of \modelName{} to baselines on the \textit{Three Block Stack}, \textit{Three-Piece Assembly}, and \textit{Coffee} Preparation tasks. Additionally, we perform ablations by removing components of \modelName{} to assess their individual contributions.





\modelName{} consistently outperforms baseline methods across all three tasks. The performance gap is largest for \textit{Three-Piece Assembly}, where \modelName{} nearly doubles the performance of MimicGen. 
In the \textit{Three Block Stack} task, \modelName{} benefits from equivariant, causal, and visual augmentation in a compositional manner. Note that we omit color jitter from \modelName{} for this experiment since the task is causally dependent on the color of the blocks. 

This suggests that counterfactual augmentation benefits most from complex tasks consisting of multiple substeps. This is consistent with the fact that in the \textit{Coffee} task, performance is roughly even amongst all baselines and method variants utilizing equivariant data augmentation. \modelName{} does not seem to benefit from visual augmentation for these tasks, however. This may be because this task demands precision, and there is a trade-off between robustness and accuracy, as demonstrated in \cite{lechner2023revisiting}. In essence, because visual augmentation demands that the BC policy be robust to a span of translations in the camera plane, image scales, and color variations, the model has less capacity to fit the action distribution, and thus the policy loses fidelity.



We further investigate the performance of individual augmentations on the transport environment (Table \ref{tab:transportaug}). 
We show that counterfactual data augmentation and random resize and crop achieve $71\%$ for single camera view and and $94\%$ accuracy for multi-view cameras, outperforming other augmentations. 
Notably, in simpler tasks, visual augmentations like resize and crop may yield high success rates because they provide sufficient variability for training robust policies. 
We find, however, that counterfactual data augmentation performs best in complex, long-horizon tasks due to its ability to maintain causal relationships between states and actions, and marginalize over subsets of the state space that are causally invariant to actions.



\begin{table}[!t]
\centering
\scriptsize 
\resizebox{0.9\columnwidth}{!}{
\begin{tabular}{l|c|c} \hline
    \toprule
    \rowcolor[HTML]{CBCEFB} 
    Augmentation Type& Single View & Multi-view \\ 
    \midrule
    No Augmentation & 39 &79\\ 
    \rowcolor[HTML]{EFEFEF} 
    Channel Permutation & 56&92\\ 
    Color Jitter & 45 &83\\ 
    \rowcolor[HTML]{EFEFEF} 
    Priprioception Noise &31 & 84\\ 
    Random Resize \& Crop &\textbf{71}& 90\\ 
    
    \rowcolor[HTML]{EFEFEF} 
    Counterfactual Augmentation &\textbf{71}& \textbf{94}\\ 
    \bottomrule
\end{tabular}
}
\caption{ACT policy performance averaged over three seeds (\%) on \texttt{Transport} when trained with various augmentations. While vision-based augmentation aids in generalization they are not considerate of causal relationships. Augmentation can be detrimental to performance if causality is not taken into consideration (e.g. color augmentation on \texttt{Three Block Stack}).}
\label{tab:transportaug}
\vspace{-3mm}
\end{table}


\paragraph{\textbf{2. Generalization}}
To address \textbf{RQ2}, we evaluate the generalization ability of \modelName{} on a Libero-Object task \cite{liu2024libero}. In particular, we test the policy's ability to generalize to:

\begin{itemize}
    \item \textbf{Unseen Distractors:} Training with a subset (2/5) of distractor objects and testing with the whole set to assess robustness to irrelevant objects in the environment.
    \item \textbf{Unseen Textures:} Applying unseen textures to objects and backgrounds at test time. Random colors are applied over objects while preserving details such as opacity, text, etc.
\end{itemize}


We test several models in these evaluation settings (Table \ref{tab:generalization}). We select one task from the Libero Object dataset ("pick up the tomato sauce and place it in the basket"). All models are trained on a source dataset of just 10 demonstrations. These demonstrations are expanded using $SE(3)$ equivariance \cite{mandlekar2023mimicgen} to 200 demonstrations. We apply causal augmentation and image augmentations (Resize/Crop and Color Jitter) to the expanded dataset. We evaluate the policy under the conditions outlined above to test generalization.
Notably, \modelName{} matches or exceeds the performance of MimicGen~\cite{mandlekar2023mimicgen}, and consistently outperforms other popular forms of data augmentation, suggesting that \modelName{} enables robustness to changes in the environment. 

\paragraph{\textbf{3. Scaling Synthetic Data}}%
To address \textbf{RQ3}, we evaluate the performance of \modelName{} when trained on varying amounts of demonstrations. We scale the number of demonstrations using $SE(3)$ equivariance and quantify the effect of diversity and complexity in the dataset (Table \ref{tab:scaling}).

\begin{table}[!t]
\centering
\scriptsize 
\resizebox{\columnwidth}{!}{
\begin{tabular}{l|c|c|c} 
    \toprule
    \rowcolor[HTML]{CBCEFB} 
    & In Distribution & OOD: Texture & OOD: Distractors \\ 
    \midrule
    \rowcolor[HTML]{EFEFEF}
    ACT (vanilla) & 76.3\,\textpm\,3.51 &  64.0\,\textpm\,8.89 & 22.3\,\textpm\,1.15 \\ 
    Resize/Crop & 46.7\,\textpm\,4.51  & 53.0\,\textpm\,5.20 & 81.7\,\textpm\,4.73 \\ 
    \rowcolor[HTML]{EFEFEF}
    Channel Permutation & 3.0\,\textpm\,1.00  & 7.7\,\textpm\,2.31 & 2.3\,\textpm\,1.15 \\ 
    Gaussian Blur & 4.7\,\textpm\,2.08   & 4.3\,\textpm\,1.53 & 11.3\,\textpm\,2.52 \\ 
    \rowcolor[HTML]{EFEFEF}
    Proprioceptive Noise & 2.7,\textpm\,0.58 & 3.3\,\textpm\,1.53 & 18.3\,\textpm\,3.21 \\ 
    Color Jitter & 96.0\,\textpm\,1.73 &  83.7\,\textpm\,4.16 & 81.3\,\textpm\,2.08 \\ 
    \rowcolor[HTML]{EFEFEF}
    CoDA~\cite{pitis2020counterfactual} & 10.7\,\textpm\,3.06 &  6.3\,\textpm\,4.04 & 7.3\,\textpm\,3.79 \\
    MimicGen \cite{mandlekar2023mimicgen} & 99.0\,\textpm\,0.00 & 99.0\,\textpm\,0.00 & \textbf{99.0\,\textpm\,0.00} \\ 
    \rowcolor[HTML]{EFEFEF}
    \modelName{} (Ours) (no Visual) & \textbf{100.0\,\textpm\,0.00}  & \textbf{100.0\,\textpm\,0.00} & \textbf{99.3\,\textpm\,0.58} \\
    \modelName{} (Ours) & \textbf{100.0\,\textpm\,0.00}  & \textbf{100.0\,\textpm\,0.00} & \textbf{99.3\,\textpm\,0.58} \\
    \rowcolor[HTML]{EFEFEF}
    
    \bottomrule
    
\end{tabular}
}
\caption{Generalization experiment on Libero Object. \modelName{} matches or exceeds the performance of other forms of data augmentation at generalizing to unseen distributions. \textit{In Distribution} refers to evaluation on the training distribution.}
\label{tab:generalization}
\vspace{-3mm}
\end{table}

We compare the performance of models trained with and without causal-based data augmentation across the \textit{Coffee} and \textit{Three Block Stack} tasks with varying number of demonstrations (10, 50, 100, 200, and 1000). For the \textit{Three Block Stack} task, counterfactual data augmentation on 200 demonstrations improves performance to $30\%$ success compared to a lesser number of demonstrations, suggesting augmentation performance scales and continues to improve at 1000 demonstrations, but at a lesser rate.

In contrast, the \textit{Coffee} task, while also requiring precise actions, may have a simpler causal complexity. The number of possible combinations of independent state partitions and causal phases is smaller in comparison to the \textit{Three Block Stack} task. This suggests that in simpler tasks, causal augmentation is less advantageous but still offers benefits by introducing diversity.

\subsection{Emergent capabilities}
We identify emergent behaviors that the policy was not explicitly trained to exhibit. One such behavior is \textbf{re-grasping}, where the policy recovers from a failed grasp attempt by reopening the gripper and attempting to grasp again (Figure \ref{fig:case_study}). We observed in rollouts that once the gripper is fully closed, the robot fails to continue the task as this action is not in the training distribution (grippers are only closed partially when picking up objects). During sequences where the robot is in a transition phase and not interacting with other entities, local factorization of the gripper position and augmenting the states of the gripper positions can be performed. This learned behavior indicates a level of adaptability and error recovery not present in the training data.

\begin{table}[!t]
\centering
\scriptsize 
\resizebox{\columnwidth}{!}{
\begin{tabular}{l|cc|cc} 
    \toprule 
    \rowcolor[HTML]{CBCEFB} 
    & \multicolumn{2}{c|}{Coffee} & \multicolumn{2}{c}{Three Block Stack} \\ 
    \midrule
    \rowcolor[HTML]{EFEFEF}
    & w/ Causal & w/o Causal & w/ Causal & w/o Causal \\
    \midrule
    10 demos &0 & 0 & 0&0\\ 
    \rowcolor[HTML]{EFEFEF}
    50 demos &4 & 9& 0&0 \\ 
    100 demos &12 & 9 & 0& 1\\ 
    \rowcolor[HTML]{EFEFEF} 
    200 demos &23 & 21 & 30 &10\\ 
    1000 demos & 50 & 47 & 70 & 63\\ 
    \bottomrule
\end{tabular}
}
\caption{Data scaling and the impact of a diverse dataset (\%). Counterfactual data augmentation excels in complex environments with multiple sub-tasks. \ag{ok so what is the big picture take away} \jeremy{TODO: Bottom row was incorrect in original submission, verify that new numbers are correct}}

\label{tab:scaling}
\vspace{-3mm}
\end{table}

\section{Conclusion}

\modelName{} provides a principled and effective framework for data augmentation in imitation learning that leverages causal invariance, geometric equivariance, and visual symmetries. This unified perspective advances the theoretical understanding of policy learning and offers practical benefits for developing robust and generalizable robotic systems. Policies trained with \modelName{} generally exhibited higher task success rates, robust generalization to unseen object poses, textures, and distractors, and improved sample efficiency, requiring fewer demonstrations to achieve comparable or superior performance.








\clearpage
\newpage
\renewcommand*{\bibfont}{\small}
\bibliographystyle{IEEEtran.bst}
\bibliography{causal-dataaug}

\begin{thebibliography}{10}
\providecommand{\url}[1]{#1}
\csname url@rmstyle\endcsname
\providecommand{\newblock}{\relax}
\providecommand{\bibinfo}[2]{#2}
\providecommand\BIBentrySTDinterwordspacing{\spaceskip=0pt\relax}
\providecommand\BIBentryALTinterwordstretchfactor{4}
\providecommand\BIBentryALTinterwordspacing{\spaceskip=\fontdimen2\font plus
\BIBentryALTinterwordstretchfactor\fontdimen3\font minus \fontdimen4\font\relax}
\providecommand\BIBforeignlanguage[2]{{%
\expandafter\ifx\csname l@#1\endcsname\relax
\typeout{** WARNING: IEEEtran.bst: No hyphenation pattern has been}%
\typeout{** loaded for the language `#1'. Using the pattern for}%
\typeout{** the default language instead.}%
\else
\language=\csname l@#1\endcsname
\fi
#2}}

\bibitem{goyal2023rvt}
A.~Goyal, J.~Xu, Y.~Guo, V.~Blukis, Y.-W. Chao, and D.~Fox, ``Rvt: Robotic view transformer for 3d object manipulation,'' in \emph{Conference on Robot Learning}.\hskip 1em plus 0.5em minus 0.4em\relax PMLR, 2023.

\bibitem{zhao2023learning}
T.~Z. Zhao, V.~Kumar, S.~Levine, and C.~Finn, ``Learning fine-grained bimanual manipulation with low-cost hardware,'' \emph{arXiv preprint arXiv:2304.13705}, 2023.

\bibitem{chi2023diffusion}
C.~Chi, S.~Feng, Y.~Du, Z.~Xu, E.~Cousineau, B.~Burchfiel, and S.~Song, ``Diffusion policy: Visuomotor policy learning via action diffusion,'' \emph{arXiv preprint arXiv:2303.04137}, 2023.

\bibitem{lee2024behavior}
S.~Lee, Y.~Wang, H.~Etukuru, H.~J. Kim, N.~M.~M. Shafiullah, and L.~Pinto, ``Behavior generation with latent actions,'' \emph{arXiv preprint arXiv:2403.03181}, 2024.

\bibitem{mete2024quest}
A.~Mete, H.~Xue, A.~Wilcox, Y.~Chen, and A.~Garg, ``Quest: Self-supervised skill abstractions for learning continuous control,'' \emph{arXiv preprint arXiv:2407.15840}, 2024.

\bibitem{brohan2022rt}
A.~Brohan, N.~Brown, J.~Carbajal, Y.~Chebotar, J.~Dabis, C.~Finn, K.~Gopalakrishnan, K.~Hausman, A.~Herzog, J.~Hsu, \emph{et~al.}, ``Rt-1: Robotics transformer for real-world control at scale,'' \emph{arXiv preprint arXiv:2212.06817}, 2022.

\bibitem{vuong2023open}
Q.~Vuong, S.~Levine, H.~R. Walke, K.~Pertsch, A.~Singh, R.~Doshi, C.~Xu, J.~Luo, L.~Tan, D.~Shah, \emph{et~al.}, ``Open x-embodiment: Robotic learning datasets and rt-x models,'' in \emph{Towards Generalist Robots: Learning Paradigms for Scalable Skill Acquisition@ CoRL2023}, 2023.

\bibitem{khazatsky2024droid}
A.~Khazatsky, K.~Pertsch, S.~Nair, A.~Balakrishna, S.~Dasari, S.~Karamcheti, S.~Nasiriany, M.~K. Srirama, L.~Y. Chen, K.~Ellis, \emph{et~al.}, ``Droid: A large-scale in-the-wild robot manipulation dataset,'' \emph{arXiv preprint arXiv:2403.12945}, 2024.

\bibitem{walke2023bridgedata}
H.~R. Walke, K.~Black, T.~Z. Zhao, Q.~Vuong, C.~Zheng, P.~Hansen-Estruch, A.~W. He, V.~Myers, M.~J. Kim, M.~Du, \emph{et~al.}, ``Bridgedata v2: A dataset for robot learning at scale,'' in \emph{Conference on Robot Learning}.\hskip 1em plus 0.5em minus 0.4em\relax PMLR, 2023.

\bibitem{shridhar2023perceiver}
M.~Shridhar, L.~Manuelli, and D.~Fox, ``Perceiver-actor: A multi-task transformer for robotic manipulation,'' in \emph{Conference on Robot Learning}.\hskip 1em plus 0.5em minus 0.4em\relax PMLR, 2023.

\bibitem{wang2023affordance}
L.~Wang, N.~Dvornik, R.~Dubeau, M.~Mittal, and A.~Garg, ``Self-supervised learning of action affordances as interaction modes,'' in \emph{2023 IEEE International Conference on Robotics and Automation (ICRA)}, 2023, pp. 7279--7286.

\bibitem{laskin2020RLwithAugData}
M.~Laskin, K.~Lee, A.~Stooke, L.~Pinto, P.~Abbeel, and A.~Srinivas, ``Reinforcement learning with augmented data,'' \emph{CoRR}, vol. abs/2004.14990, 2020.

\bibitem{young2020VIeasy}
S.~Young, D.~Gandhi, S.~Tulsiani, A.~Gupta, P.~Abbeel, and L.~Pinto, ``Visual imitation made easy,'' \emph{CoRR}, vol. abs/2008.04899, 2020.

\bibitem{mandlekar2023mimicgen}
A.~Mandlekar, S.~Nasiriany, B.~Wen, I.~Akinola, Y.~Narang, L.~Fan, Y.~Zhu, and D.~Fox, ``Mimicgen: A data generation system for scalable robot learning using human demonstrations,'' \emph{arXiv preprint arXiv:2310.17596}, 2023.

\bibitem{kostrikov2020augallyouneed}
I.~Kostrikov, D.~Yarats, and R.~Fergus, ``Image augmentation is all you need: Regularizing deep reinforcement learning from pixels,'' \emph{CoRR}, vol. abs/2004.13649, 2020.

\bibitem{mandlekar2021matters}
A.~Mandlekar, D.~Xu, J.~Wong, S.~Nasiriany, C.~Wang, R.~Kulkarni, L.~Fei-Fei, S.~Savarese, Y.~Zhu, and R.~Mart{\'\i}n-Mart{\'\i}n, ``What matters in learning from offline human demonstrations for robot manipulation,'' \emph{arXiv preprint arXiv:2108.03298}, 2021.

\bibitem{akkaya2019solving}
I.~Akkaya, M.~Andrychowicz, M.~Chociej, M.~Litwin, B.~McGrew, A.~Petron, A.~Paino, M.~Plappert, G.~Powell, R.~Ribas, \emph{et~al.}, ``Solving rubik's cube with a robot hand,'' \emph{arXiv preprint arXiv:1910.07113}, 2019.

\bibitem{sinha2022s4rl}
S.~Sinha, A.~Mandlekar, and A.~Garg, ``S4rl: Surprisingly simple self-supervision for offline reinforcement learning in robotics,'' in \emph{Conference on Robot Learning}.\hskip 1em plus 0.5em minus 0.4em\relax PMLR, 2022.

\bibitem{chen2023genaug}
Z.~Chen, S.~Kiami, A.~Gupta, and V.~Kumar, ``Genaug: Retargeting behaviors to unseen situations via generative augmentation,'' \emph{arXiv preprint arXiv:2302.06671}, 2023.

\bibitem{yu2023scaling}
T.~Yu, T.~Xiao, A.~Stone, J.~Tompson, A.~Brohan, S.~Wang, J.~Singh, C.~Tan, J.~Peralta, B.~Ichter, \emph{et~al.}, ``Scaling robot learning with semantically imagined experience,'' \emph{arXiv preprint arXiv:2302.11550}, 2023.

\bibitem{lu2024synthetic}
C.~Lu, P.~Ball, Y.~W. Teh, and J.~Parker-Holder, ``Synthetic experience replay,'' \emph{Advances in Neural Information Processing Systems}, vol.~36, 2024.

\bibitem{pitis2020counterfactual}
S.~Pitis, E.~Creager, and A.~Garg, ``Counterfactual data augmentation using locally factored dynamics,'' \emph{Advances in Neural Information Processing Systems}, 2020.

\bibitem{pitis2022mocoda}
S.~Pitis, E.~Creager, A.~Mandlekar, and A.~Garg, ``Mocoda: Model-based counterfactual data augmentation,'' \emph{Advances in Neural Information Processing Systems}, 2022.

\bibitem{lu2020sample}
C.~Lu, B.~Huang, K.~Wang, J.~M. Hern{\'a}ndez-Lobato, K.~Zhang, and B.~Sch{\"o}lkopf, ``Sample-efficient reinforcement learning via counterfactual-based data augmentation,'' \emph{arXiv preprint arXiv:2012.09092}, 2020.

\bibitem{brohan2023rt}
A.~Brohan, N.~Brown, J.~Carbajal, Y.~Chebotar, X.~Chen, K.~Choromanski, T.~Ding, D.~Driess, A.~Dubey, C.~Finn, \emph{et~al.}, ``Rt-2: Vision-language-action models transfer web knowledge to robotic control,'' \emph{arXiv preprint arXiv:2307.15818}, 2023.

\bibitem{bharadhwaj2024roboagent}
H.~Bharadhwaj, J.~Vakil, M.~Sharma, A.~Gupta, S.~Tulsiani, and V.~Kumar, ``Roboagent: Generalization and efficiency in robot manipulation via semantic augmentations and action chunking,'' in \emph{2024 IEEE International Conference on Robotics and Automation (ICRA)}.\hskip 1em plus 0.5em minus 0.4em\relax IEEE, 2024.

\bibitem{lechner2023revisiting}
M.~Lechner, A.~Amini, D.~Rus, and T.~A. Henzinger, ``Revisiting the adversarial robustness-accuracy tradeoff in robot learning,'' \emph{IEEE Robotics and Automation Letters}, vol.~8, no.~3, pp. 1595--1602, 2023.

\bibitem{liu2024libero}
B.~Liu, Y.~Zhu, C.~Gao, Y.~Feng, Q.~Liu, Y.~Zhu, and P.~Stone, ``Libero: Benchmarking knowledge transfer for lifelong robot learning,'' \emph{Advances in Neural Information Processing Systems}, vol.~36, 2024.

\end{thebibliography}
\end{document}